\documentclass[lettersize,journal]{IEEEtran}
\usepackage{amsmath,amsfonts}
\usepackage{algorithmic}
\usepackage{algorithm}
\usepackage{array}
\usepackage[caption=false,font=normalsize,labelfont=sf,textfont=sf]{subfig}
\usepackage{textcomp}
\usepackage{stfloats}
\usepackage{url}
\usepackage{verbatim}
\usepackage{graphicx}
\usepackage{cite}

\usepackage{bm}
\usepackage{amssymb}

\usepackage{enumitem}
\usepackage[bookmarks=true, hidelinks]{hyperref}

\usepackage{multirow}
\usepackage{multicol}
\usepackage{makecell}

\usepackage{amsmath}
\usepackage{amssymb}
\usepackage{mathtools} % for prescript
\usepackage{bm}

\newcommand{\posintset}[2]{\mathbb{N}_{[{#1},{#2}]}}
\newcommand{\basevar}[4]{\prescript{#4}{}{#1}^{#3}_{#2}}

\newcommand{\rstate}[1]{\basevar{\bm{s}}{#1}{}{}}
\newcommand{\rstatei}[2]{\basevar{\bm{s}}{#1}{(#2)}{}}
\newcommand{\rstateref}[1]{\basevar{\Tilde{\bm{s}}}{#1}{}{}}

\newcommand{\raction}[1]{\basevar{\bm{u}}{#1}{}{}}
\newcommand{\ractioni}[2]{\basevar{\bm{u}}{#1}{(#2)}{}}
\newcommand{\ractionref}[1]{\basevar{\Tilde{\bm{u}}}{#1}{}{}}

\newcommand{\rpos}[1]{\basevar{\bm{p}}{#1}{}{}}
\newcommand{\rposi}[2]{\basevar{\bm{p}}{#1}{(#2)}{}}
\newcommand{\stcobsset}[0]{\mathcal{O}}
\newcommand{\dynobsset}[1]{\mathcal{D}_{#1}}
\newcommand{\dynobssetest}[1]{\Hat{\mathcal{D}}_{#1}}

\newcommand{\hstatei}[3]{\basevar{\bm{z}}{#1}{(#2)}{#3}}

\newcommand{\hstateiest}[3]{\basevar{\Hat{\bm{z}}}{#1}{(#2)}{#3}}
\newcommand{\hposi}[3]{\basevar{\bm{q}}{#1}{(#2)}{#3}}

\begin{document}

\title{Future-Oriented Navigation: Dynamic Obstacle Avoidance with One-Shot Energy-Based Multimodal Motion Prediction}

\author{
Ze Zhang, %~\IEEEmembership{Graduate Student Member,~IEEE}, 
Georg Hess, %~\IEEEmembership{Graduate Student Member,~IEEE}, 
Junjie Hu, \\
Emmanuel Dean, 
Lennart Svensson, %~\IEEEmembership{Senior Member,~IEEE} 
and Knut Åkesson,%~\IEEEmembership{Member,~IEEE}
        % <-this % stops a space
\thanks{Manuscript received September 30, 2024; revised April 27, 2025 and February 17, 2025; Accepted: May 21, 2025.}
\thanks{This paper was recommended for publication by Gentiane Venture upon evaluation of the Associate Editor and Reviewers’ comments.}
\thanks{This work is supported by the AIHURO project (Vinnova 2022-03012). The computations were enabled by resources provided by the National Academic Infrastructure for Supercomputing in Sweden (NAISS), partially funded by the Swedish Research Council (grant agreement no. 2022-06725).}% <-this % stops a space
\thanks{All authors are with Chalmers University of Technology, 41296 Gothenburg, Sweden ({\tt\scriptsize \{zhze, georghe, junjieh, deane, lennart.svensson, knut\}@chalmers.se}).}
\thanks{Digital Object Identifier (DOI): see top of this page.}
}

% The paper headers
\markboth{IEEE ROBOTICS AND AUTOMATION LETTERS. PREPRINT VERSION. ACCEPTED May, 2025}%
{Zhang \MakeLowercase{\textit{et al.}}: Prescient Collision-Free Navigation of Mobile Robots with Iterative MMP of Dynamic Obstacles}

% \IEEEpubid{0000--0000/00\$00.00~\copyright~2021 IEEE}
% Remember, if you use this you must call \IEEEpubidadjcol in the second
% column for its text to clear the IEEEpubid mark.

\maketitle

\begin{abstract}
This paper proposes an integrated approach for the safe and efficient control of mobile robots in dynamic and uncertain environments. The approach consists of two key steps: one-shot multimodal motion prediction to anticipate motions of dynamic obstacles and model predictive control to incorporate these predictions into the motion planning process.
Motion prediction is driven by an energy-based neural network that generates high-resolution, multi-step predictions in a single operation. The prediction outcomes are further utilized to create geometric shapes formulated as mathematical constraints. Instead of treating each dynamic obstacle individually, predicted obstacles are grouped by proximity in an unsupervised way to improve performance and efficiency. The overall collision-free navigation is handled by model predictive control with a specific design for proactive dynamic obstacle avoidance. The proposed approach allows mobile robots to navigate effectively in dynamic environments. Its performance is accessed across various scenarios that represent typical warehouse settings. The results demonstrate that the proposed approach outperforms other existing dynamic obstacle avoidance methods.
\end{abstract}

\begin{IEEEkeywords}
Human-aware motion planning, collision avoidance, deep learning methods.
\end{IEEEkeywords}

\section{Introduction}
\IEEEPARstart{F}{actory}, warehouse, and laboratory layouts are subject to changes, and the presence of humans and manually driven vehicles further complicates motion planning for Autonomous Mobile Robots (AMRs). 
To ensure collision avoidance, AMRs must anticipate interactions with static and dynamic objects, though the unpredictability of humans poses significant challenges. 
A prevalent strategy is to halt the robot when obstacles are detected \cite{agv_2020}, risking downtime if obstructions persist or deadlocks occur. 
This study addresses collision-free navigation for AMRs in industrial settings but can be generalized to other settings with similar traffic dynamics.

Dynamic Obstacle Avoidance (DOA) has been a key research area since the advent of AMRs, with an initial emphasis on safety. This often leads to overly cautious behavior, exemplified by the Freezing Robot Problem (FRP) \cite{frp_2010}, where the planner deems all paths and stops the robot.
Advancements in perception and computation technologies enable robots to rapidly and accurately gather extensive environmental data and coordinate actions, improving the ability to implement effective decision-making strategies.
Recent research has improved motion planning efficiency and addressed FRP by incorporating human movements and intentions \cite{unfrozen_2020, attentionbased_2023}, especially in social robot navigation \cite{socialnavi_2023}. In traditional model-based approaches, the Constant Velocity Model (CVM) and velocity-based strategies are widely used for predicting human motion in navigation frameworks \cite{unfrozen_2020, dynchannel_2019, dynsys_2022}. Additionally, predefined kinematics-based \cite{due_2012} and interaction-based \cite{proactive_2023} models offer more detailed insights for motion planners. 
Recently, learning-based models have become the primary motion predictor in most prescient navigation frameworks due to their high flexibility and ability to interpret complex and uncertain interactions and environments \cite{proactive_2023, attentionbased_2023, groupbased_2022, ze_2023}.

\begin{figure}[t]
    \centering
    \includegraphics[width=0.9\linewidth]{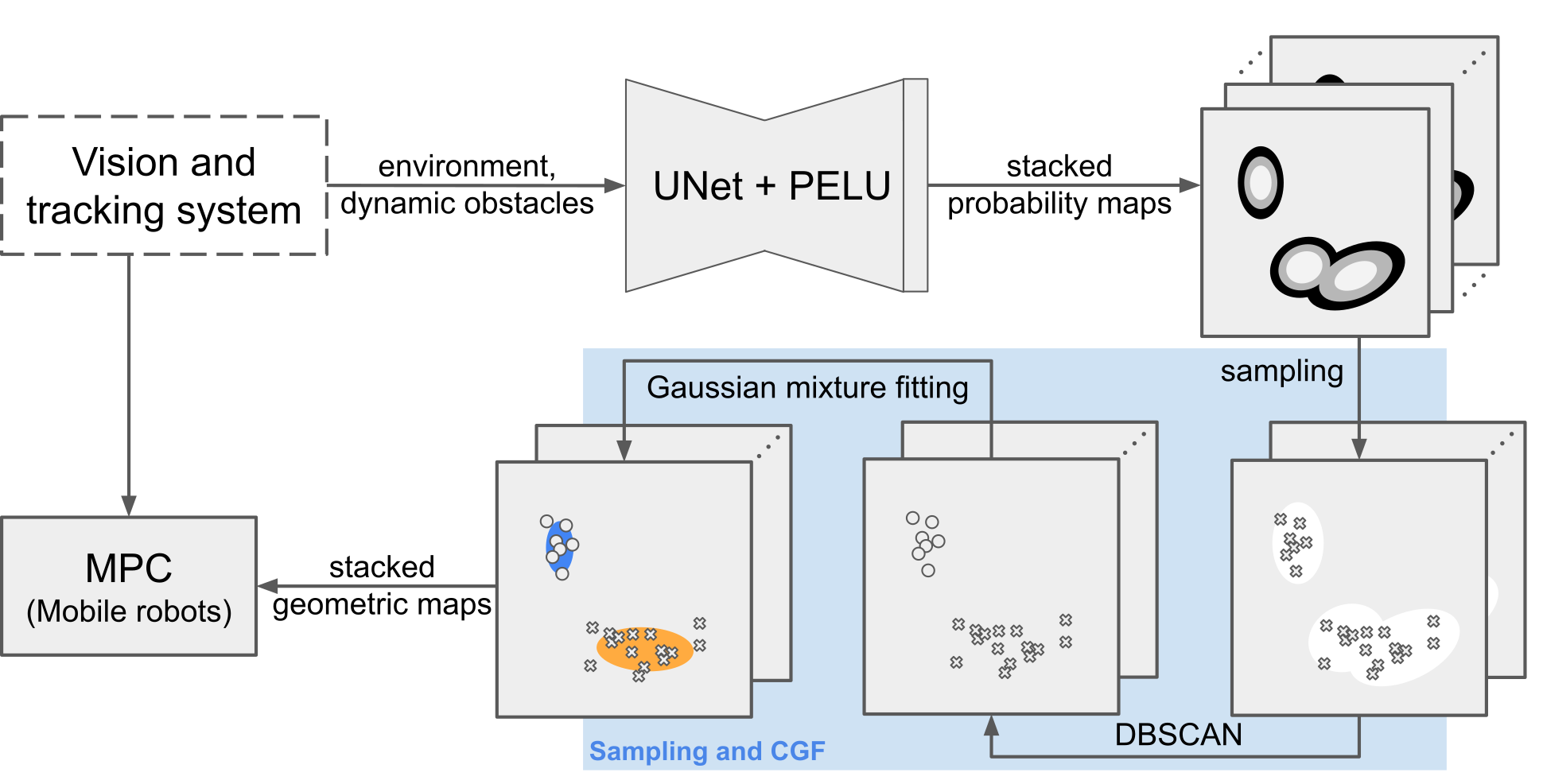}
    \vspace{-3mm}
    \caption{Pipeline of the proposed obstacle avoidance approach, assuming environmental information is provided by a bird's-eye-view vision system.}
    \label{fig:pipline}
    \vspace{-3mm}
\end{figure}

This work presents a navigation framework for multiple AMRs in dynamic environments, tailored for industrial transport tasks, incorporating multimodal motion prediction, prediction grouping, static obstacle avoidance, proactive DOA, and fleet collision avoidance. 
The approach integrates a data-driven motion predictor with a predictive trajectory planner. Specifically, an energy-based model predicts possible future motions of dynamic obstacles as probability maps, which are used by distributed Model Predictive Control (MPC) for safe navigation. 
Multiple tests are designed to evaluate the method's performance in a comparative study with state-of-the-art approaches. The main contributions are threefold:
\begin{itemize}[leftmargin=*]
    \item Introducing a fast and stable energy-based learning approach for multimodal motion prediction enabling improved integration with downstream planning and control.
    \item Improving MPC-based motion planner for effective, safe, and efficient fleet control in dynamic environments.
    \item Evaluating the framework under various conditions and scenarios, with and without cooperative pedestrians.
\end{itemize}

\section{Related Work}
% Multimodal Motion Prediction
Motion prediction of dynamic obstacles is crucial for the downstream motion planning of AMRs. Early CVM-based methods rely solely on velocity information and have been proven effective \cite{cvm_2020}, but struggle with human behavior's uncertainty. Other model-based approaches, such as social force \cite{sf_1995}, reciprocal velocity obstacles \cite{rvo_2008}, and other kinematic models \cite{due_2012}, also fail to capture the multimodality required to address this uncertainty. 
Advancements in deep learning have enabled the effective capture of environmental contexts and agent interactions while supporting uncertainty and multimodality. Techniques such as social pooling \cite{slstm_2016_CVPR, sgan_2018_CVPR} model pedestrian interactions, Convolutional Neural Networks (CNNs) enhance context perception \cite{overcoming_2019, ynet_2021, mpssd_2020}, attention mechanisms \cite{trajectronpp_2020} provide a comprehensive and global understanding, and inverse reinforcement learning \cite{end2end_2022} encodes environmental information as rewards. 

Reinforcement learning has been applied to AMR motion planning and obstacle avoidance \cite{unfrozen_2020, ddpg_2024}, providing flexibility but facing instability and safety challenges. Field-based methods \cite{dynsys_2022} have been developed to generate smooth motion fields around obstacles. However, these methods fail to address the inherent uncertainty in the environment.
Industrial transport robots require stability, flexibility, interpretability, and predictive capabilities. As an optimization-based method, MPC is well-suited for DOA due to its constraint handling and predictive receding horizon. Combining MPC with deep learning for motion prediction improves accuracy and effectiveness in dynamic environments \cite{proactive_2023, groupbased_2022, ze_2021, ze_2023}.

This work follows a similar modular structure as in \cite{ze_2023} but with two key differences, including a faster and more stable neural network for multimodal motion prediction and an extended MPC controller to achieve fleet collision avoidance \cite{dist_2022}. A novel Energy-Based Learning (EBL) strategy is proposed to apply the new neural network architecture.

\section{PROBLEM FORMULATION}
This work addresses two coherent challenges: AMR controller design for DOA and motion prediction of dynamic obstacles, focusing on 2D spaces and discrete-time domains.

\noindent\textbf{Collision-free Motion Planning}:
Let the discrete motion model for the $i$-th AMR is $\rstatei{k+1}{i}=f(\rstatei{k}{i}, \ractioni{k}{i})$ in a fleet, where $\rstatei{k}{i}$ and $\ractioni{k}{i}$ represent its state and action at time step $k$. The aim is to plan its trajectory along a given reference path while avoiding collisions with obstacles or other robots and adhering to its physical limitations. For simplicity, the identification index $(i)$ is omitted when discussing a single robot for all variables.
Static obstacles are assumed to be polygonal. Their occupied area, $\stcobsset=\cup_j\stcobsset^{(j)}$, is the union of individuals, where each set is a closed intersection of half-spaces.
Dynamic obstacles are modeled as ellipses and the corresponding area $\dynobsset{k}$ may vary over time. Similarly, $\dynobsset{k}=\cup_j\dynobsset{k}^{(j)}$. At any time, a robot at position $\rpos{k}$ must be outside any obstacles: $\forall k, \rpos{k}\notin\stcobsset\cup\dynobsset{k}$. Obstacles are inflated by the size of the robot, and robots are regarded as points \cite{ze_2023}.

\noindent\textbf{Motion Prediction}:
To consider potential collisions with dynamic obstacles, AMRs should have access to the estimated future positions of obstacles. Let the predictive horizon be $N$, AMRs require information on other dynamic objects from the current instant to future $N$ steps. For dynamic obstacle $i$, the task is to predict its future states $\{\hstatei{t_k}{i}{}\}^{k+N}_{{t_k}=k}$. Since future motion can be multimodal, multiple hypotheses of each future position should be available. For a maximal number $M$ of modes at each time step for each dynamic obstacle, the desired prediction is notated as $\{\{\hstatei{t_k}{i}{m}\}^M_{m=1}\}^{k+N}_{{t_k}=k}$.

\section{Future-Oriented Model Predictive Control}
This section outlines the formulation of the MPC problem incorporating DOA with consideration of multiple futures.

\subsection{Dynamic Obstacle Avoidance with Multiple Futures}
A dynamic obstacle with $M$ possible futures can be viewed as $M$ potential obstacles, each with a different probability of occurrence. Assuming accurate predictions, a sufficient condition for an AMR to avoid the obstacle within the horizon is that it avoids collisions with all potential obstacles at the corresponding time steps, as in Eq. \eqref{eq:dynobs_hc}, where $\prescript{m}{}{\dynobssetest{t_k}}$ is the predicted occupied area of obstacle $m$ at $t_k$. A positive integer set from $a$ to $b$ is denoted as $\posintset{a}{b}$.
\begin{equation}
    \forall t_k\in\posintset{k}{k+N},\ m\in\posintset{1}{M},\ \bm{p}_{t_k}\notin \prescript{m}{}{\dynobssetest{t_k}}. \label{eq:dynobs_hc}
\end{equation}
This condition can be overly restrictive, especially with large $M$, as potential obstacles may occupy the entire drivable area, rendering the MPC problem infeasible.
Retaining only predicted futures with high probabilities provides limited mitigation of this issue. Thus, two solutions are introduced in this work to address this.

Firstly, rather than enforcing predictive DOA as a hard constraint, a flexible alternative is to add it into the objective as a soft constraint, as shown in Eq. \eqref{eq:dynobs_sc} for $t_k\in\posintset{k}{k+N}$. Here, function $\iota_{D}(\cdot)$ \cite{ze_2023} evaluates the distance between the robot and the potential obstacle, $\beta_{t_k}$ is the weight corresponding to time, and $\alpha_m$ is mode-dependent.
\begin{equation}
    J_{\mathcal{D}}(t_k) = \beta_{t_k} \cdot \left[\sum^M_{m=1}\alpha_m \cdot \iota_{D}(\bm{p}_{t_k}, \prescript{m}{}{\dynobsset{t_k}})\right] \label{eq:dynobs_sc}
\end{equation}
The mode weight $\alpha_m$ can be determined by the probability of each mode's occurrence, reflecting the intrinsic uncertainty. The time weight $\beta_{t_k}$ can be chosen such that a larger $t_k$ corresponds to a smaller weight. 
The second idea is to group predictions at each step based on their proximity \cite{ze_2021}. This strategy considers occupied areas collectively rather than focusing on individuals, which not only reduces the computation burden for MPC but can also alleviate the FRP.
\subsection{Model Predictive Control Formation}
The objective of MPC is composed of three terms, including the reference deviation term $J_R$, soft DOA term $J_{\mathcal{D}}$, and fleet-collision-avoidance term \cite{dist_2022}. For a robot at time $k$,

{\small
\begin{align}
    J_R(k) &= ||\!\rstate{k}\!-\!\rstateref{k}||^2_{\bm{Q}_s} 
    \!+ ||\!\raction{k}\!-\!\ractionref{k}||^2_{\bm{Q}_u} 
    \!+ ||\!\raction{k}\!-\!\raction{k-1}||^2_{\bm{Q}_a}, \\
    J_{\mathcal{D}}(k) &= \sum_{j=1}^{n_d}||\iota_D(\rpos{k}|\hposi{k}{j}{},\bm{\sigma}^{\prime j}_k)||^2_{\bm{Q}_D},
\end{align}
}

\noindent where $\rstateref{k}$ is the reference state, $\ractionref{k}$ is the reference action, $n_d$ is the number of dynamic obstacles, $\hposi{k}{j}{}$ is the position of obstacle $j$, $\bm{\sigma}^{\prime j}_k=\bm{\sigma}^{j}_k+r_{\text{extra}}$ is the obstacle axes based on the original one $\bm{\sigma}^{j}_k$ padded by a safe margin $r_{\text{extra}}$, and all $\bm{Q}$ variables are penalty parameters. 
For the fleet collision avoidance between the ego robot $i$ and another robot $j$,
\begin{equation}
    J_F(j) = \max\left[0, \bm{Q}_f\cdot\left(d_{\text{fleet}}-||\rposi{k}{i}-\rposi{k}{j}|| \right)^2\right],
\end{equation}
where $d_{\text{fleet}}$ is the safe distance between robots.
The overall MPC problem for mobile robot $i$ is formulated as (omit $i$ if there is no ambiguity)
\begin{align}
    \min_{\bm{u}_{k:k+N-1}} & \sum_{{t_k}=k}^{k+N-1} \bigg[ J_R(t_k) + J_{\mathcal{O}}(t_k) + J_{\mathcal{D}}(t_k) \nonumber  \\
    & + \sum_{j=1,j\ne i}^{n_r}J_F(j) \bigg],  \\
    \text{s.t.} \quad 
    & \rstate{t_k+1} = f(\rstate{t_k}, \raction{t_k}), \, \forall t_k\in\posintset{k}{k+N-1}, \\
    & \raction{t_k} \in [\bm{u}_{\text{min}}, \bm{u}_{\text{max}}], \, \forall t_k\in\posintset{k}{k+N-1}, \\
    & \Dot{\bm{u}}_{t_k} \in [\Dot{\bm{u}}_{\text{min}}, \Dot{\bm{u}}_{\text{max}}], \, \forall t_k\in\posintset{k}{k+N-1}, \\
    & \bm{p}_{t_k} \notin \mathcal{O}, \, \forall t_k\in\posintset{k}{k+N-1}, \\
    & \bm{p}_{t_k} \notin \mathcal{D}_{t_k}, \, \forall t_k\in\posintset{k}{k+N_{\text{crit}}}.
\label{eq:mpc}
\end{align}
where $n_r$ is the number of robots, $N_{\text{crit}}$ is the critical horizon used to apply hard constraints on avoiding dynamic objects, $\Dot{\bm{u}}_{t_k}$ is the derivative of the action, and $(\bm{u}_{\text{min}/\text{max}}, \Dot{\bm{u}}_{\text{min}/\text{max}})$ are physical limits of the robot's action.

\begin{figure}[t]
    \centering
    \includegraphics[width=0.99\linewidth]{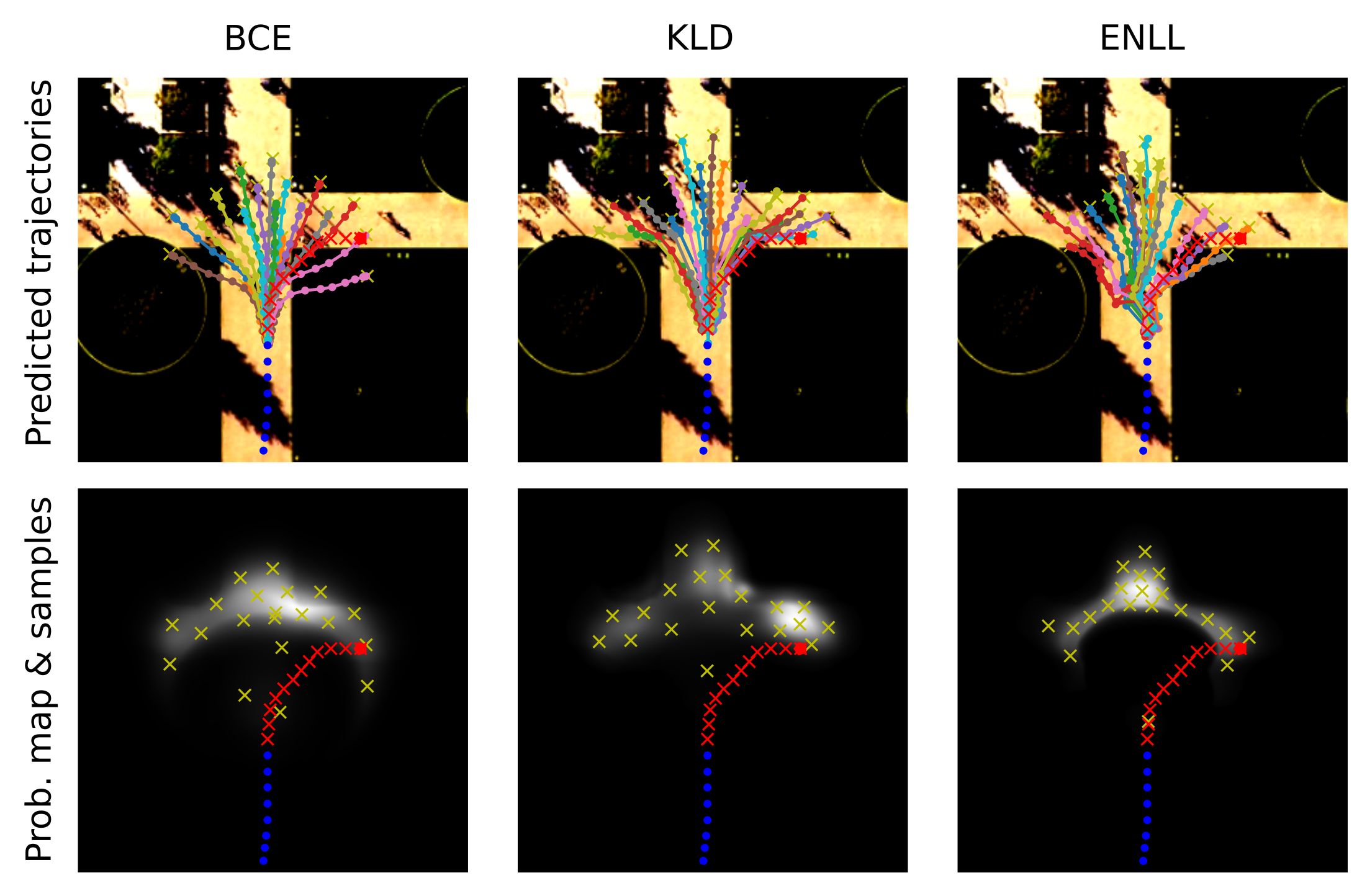}
    \vspace{-8mm}
    \caption{Training results on the Stanford Drone Dataset \cite{robicquet_2016_SDD} with different learning strategies based on the Y-Net \cite{ynet_2021} architecture. Blue dots indicate the historical trajectory of the target while red crosses are the ground-truth future trajectory. The probability maps display the predicted probabilities for the last future step. Yellow crosses are samples from the maps.}
    \label{fig:sdd_experiment}
\end{figure}

\section{Multi-step Multimodal Motion Prediction}
This section introduces a key innovation: applying energy-based learning to discrete maps for multimodal motion prediction, framed as a large-scale classification problem. It is shown that the pixel-level Negative Log-Likelihood (NLL) loss can produce concentrated predictions required for downstream planning. The main challenge with training stability is addressed by modifying the NLL loss.

\subsection{Energy-based Training}
Energy-based models \cite{lecun_2006} are models trained to yield energy spaces for given inputs. The training process is known as EBL. While commonly used in generative models, they can be applied to regression tasks \cite{ebm_2020}. 
For a model $G_\theta(\cdot)$ parameterized by $\theta$, given input $\bm{x}$ and label $\bm{y}$, it should assign a low energy to the label $\bm{y}$ and higher energy for others $\bm{y}'$, i.e., $G_\theta(\bm{x},\bm{y}) \le G_\theta(\bm{x},\bm{y}')$. 
In motion prediction, given input such as historical pedestrian motion, the model generates an energy space $\bm{E} = G_\theta(\bm{x})$ for future motion. The predicted future position $\Hat{\bm{y}}$ is obtained by querying the point in $\bm{E}$ with the lowest energy. The desired energy space should be constructed to minimize $||\bm{y}-\Hat{\bm{y}}||$. Treating $\bm{E}$ as an unnormalized probability map, multimodal motion predictions can be obtained via Monte Carlo sampling. 

\begin{figure*}[t]
    \centering
    \includegraphics[width=0.8\linewidth]{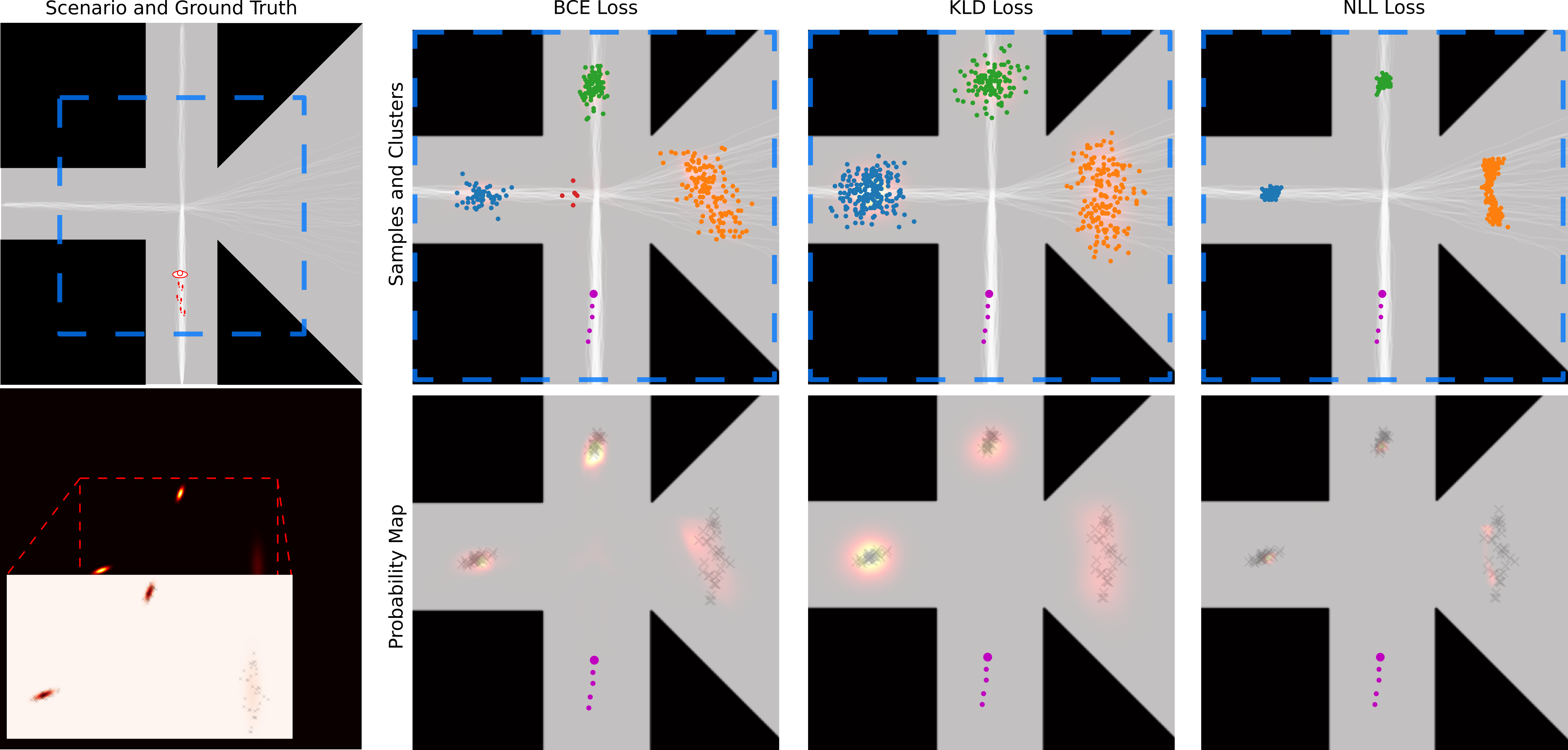}
    \vspace{-3mm}
    \caption{Comparison of training results on the synthetic crossroad dataset from different learning strategies.
    The first column shows the scenario (top) with simulated trajectories (white lines) and the estimated ground-truth probability distribution (bottom) after 20 time steps.
    For the other subfigures, purple dots indicate the current (big) and past (small) positions of the target. The first row contains predicted samples and clusters from different methods, and each color indicates a cluster.
    In the second row, black crosses on probability maps are the ground-truth future positions.}
    \label{fig:scud_experiment}
\end{figure*}

According to the Gibbs-Boltzmann distribution \cite{lecun_2006}, the energy can be converted into probability densities,
\begin{equation}
    \bm{P}_c \coloneqq \text{Pr}(\bm{y}|\bm{x};\theta) = \frac{e^{-G_\theta(\bm{x},\bm{y})}}{\int e^{-G_\theta(\bm{x},\bm{y'})}d\bm{y}'}. \label{eq:boltzman}
\end{equation}
The integral is mostly intractable \cite{ebmtraining_2021} and needs to be estimated, e.g., via importance sampling \cite{ebm_2020}. It is possible to discretize the energy space and substitute the integral by summation. For motion prediction, if the discretized grid size is equal to the resolution of the environmental image, the approximation of the integral is sufficiently accurate. In the two-dimensional situation, for a discrete energy map with width $W$ and height $H$, Eq. \eqref{eq:boltzman} can be rewritten as:
\begin{equation}
    \bm{P} \coloneqq \text{Pr}(\bm{y}|\bm{x};\theta) = \frac{e^{-\bm{E}_{w,h}}}{\sum^W_{w'=1}\sum^H_{h'=1} e^{-\bm{E}_{w',h'}}}. \label{eq:boltzman_d}
\end{equation}
where $\bm{y}=[w,h]^T$ indicating the row index $w$ and column index $h$ on the image axis. Even though Eq. \eqref{eq:boltzman_d} seems reasonable, training accordingly is unstable, and the result tends to be noisy when the size of the energy surface is large. For example, given an energy map of size $100\times100$, for each $(\bm{x},\bm{y})$, only the energy of one pixel is pushed down and the energy of the other 9999 pixels is pulled up. This makes the model sensitive to noise and likely to generate too high or low energy. To solve this problem, the weighted soft loss and a modified exponential layer \cite{elu_2016}, called the Positive Exponential Linear Unit (PELU) are proposed.

The weighted soft loss utilizes relaxed label masks. Instead of a single ground-truth pixel $\bm{y}$, a Gaussian distribution mask $\bm{A}$ centered at $\bm{y}$ with the same size of $\bm{E}$ is used. Supposing the constructed mask $\bm{A}$ to be the ground-truth distribution, which typically cannot be validated in the real world, the model can be trained to approximate this distribution via the Binary Cross Entropy (BCE) loss \cite{ynet_2021} or Kullback–Leibler Divergence (KLD). If the model is trained via BCE (which is not EBL), the direct output logits need to be processed by a Sigmoid layer and normalized into probability maps. Let $\bar{\bm{A}}$ be the normalized mask ($\bm{A}$ divided by its maximum value).
The BCE loss function can be written as
\begin{align}
    \mathcal{L}_{BCE} = 
    &-\sum_{w',h'}\left[\bar{\bm{A}}_{w',h'}\ln\sigma(\bm{E}_{w',h'})\right] \nonumber \\
    &-\sum_{w',h'}\left[(1-\bar{\bm{A}}_{w',h'})\ln(1-\sigma(\bm{E}_{w',h'}))\right].
\end{align}
where $\sigma(\cdot)$ is the element-wise Sigmoid function. Note that the notation $\bm{E}$ is kept for simplicity, but the output is not an energy space for BCE-trained models. If the model is trained via KLD (which is one form of EBL), the loss function is
\begin{align}
    \mathcal{L}_{KLD}
    = \!\sum_{w',h'}\!\bm{A}\ln\frac{\bm{A}}{\bm{P}} 
    &= \sum_{w',h'}\bm{A}_{w',h'}\bm{E}_{w',h'} \nonumber \\
    &+ \sum_{w',h'}\bm{A}_{w',h'} \ln\! \sum_{w',h'}e^{-\bm{E}_{w',h'}}.
    \label{eq:loss_kld}
\end{align}
Both BCE and KLD regard the constructed Gaussian mask as the ground truth and attempt to align the predicted distribution with it, which can lead to overestimation, as shown in Fig. \ref{fig:sdd_experiment}.
This overestimation increases the likelihood of covering the actual trajectory, particularly in open areas where dynamic objects exhibit more random movement compared to constrained areas such as factories. However, it complicates downstream motion planning by covering excessively large areas, potentially obstructing the entire drive space for AMRs.

To overcome this issue, rather than using $\bm{A}$ as a distribution, its normalized mask $\Bar{\bm{A}}$ is applied as a weight mask for the energy space to redefine the probability $\Bar{\bm{P}}$,
\begin{equation}
    \Bar{\bm{P}}_{w,h} = \frac{e^{-\Bar{\bm{A}}_{w,h}\bm{E}_{w,h}}}{\sum^W_{w'=1}\sum^H_{h'=1} e^{-\bm{E}_{w',h'}}}, \label{eq:weighted_boltzman_d}
\end{equation}
and the NLL loss is applied,
\begin{align}
    \mathcal{L}_{NLL} 
    = -\ln \Bar{\bm{P}} = 
    &-\ln\sum_{w',h'}\Bar{\bm{A}}_{w',h'}e^{-\bm{E}_{w',h'}} \nonumber \\
    &+ \ln\sum_{w',h'}e^{-\bm{E}_{w',h'}}. \label{eq:loss_nll}
\end{align}
From our training results, as shown in Fig. \ref{fig:sdd_experiment}, NLL generates more concentrated results compared to BCE and KLD.

\begin{figure*}[t]
    \centering
    \includegraphics[width=0.99\linewidth]{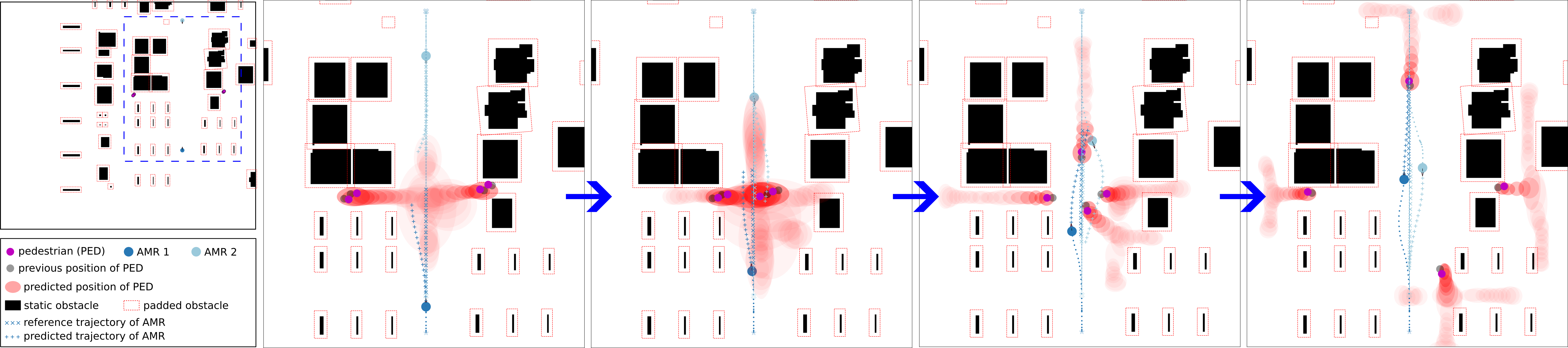}
    \vspace{-3mm}
    \caption{Scenario 3 in obstacle avoidance evaluation. Four pedestrians and two AMRs are crossing an intersection, which causes a busy area and is challenging for AMRs. The illustrated collision-free example is from MPC-ENLL. The subfigures are in chronological order from left to right.}
    \label{fig:s3}
\end{figure*}

As an EBL method, the NLL loss aggressively shapes the energy space, pushing the energy at the ground-truth cells down to a large negative value. A large negative energy leads $e^{-E}$ to infinity and thus fails the training. Our solution is to add an output layer for energy regulation by substituting the exploding part of exponentiation with a linear function, which is similar to the ELU activation function \cite{elu_2016},
\begin{equation}
    f_\text{PELU}(x) = \max(0,x) + \min(0,e^x-1) + 1 + \epsilon, \label{eq:pelu}
\end{equation}
where $\epsilon$ is a small positive offset to prevent the output from reaching zero, ensuring the logarithm remains finite. Since the output layer only produces positive values, it is called Positive ELU (PELU).
The PELU layer generates a processed energy map $\bm{E}'=\text{PELU}(-\bm{E})$, which can be converted into probability maps $\bm{P}$ according to Eq. \eqref{eq:boltzman_d},
\begin{equation}
    \bm{P}_{w,h} = \frac{\bm{E}'_{w,h}}{\sum^W_{w'=1}\sum^H_{h'=1}\bm{E}'_{w',h'}}. \label{eq:e2p}
\end{equation}
By combining Eq. \eqref{eq:loss_nll} and \eqref{eq:e2p}, the final Energy-oriented NLL (ENLL) loss function is
\begin{align}
    \mathcal{L}_\text{ENLL} = 
    - \ln\sum_{w',h'}\Bar{\bm{A}}_{w',h'}\bm{E}'_{w',h'}
    + \ln\sum_{w',h'}\bm{E}'_{w',h'}.
    \label{eq:enll}
\end{align}

\subsection{Clustering and Gaussian Fitting with Probabilities}
Samples can be drawn at each time offset on the probability map. By clustering these samples, different modes are represented by distinct clusters. Gaussian fitting is then applied to the samples in each cluster, resulting in multimodal elliptical Gaussian estimations of the future position. Alternatively, samples can be directly fit into a Gaussian Mixture Model (GMM). This process, called Clustering and Gaussian Fitting (CGF) \cite{ze_2021}, yields the estimated future states $\{\{\hstateiest{t_k}{i}{m}\}^M_{m=1}\}^{k+N}_{{t_k}=k}$ of dynamic obstacle areas $\{\dynobssetest{t_k}\}^{k+N}_{t_k=k}$.
Since each sample has a probability density, the probability of a cluster can be estimated by either the maximum or the weighted average probability of the samples in that cluster. If a GMM is directly obtained, the component weights can serve as the cluster probabilities.

\begin{table}[t]
    \centering
    \caption{Quantitative comparison of different methods in Fig. \ref{fig:scud_experiment}.}
    \vspace{-3mm}
    \begin{tabular}{|c|c|c|c|}
    \hline
        Metrics (lower is better) & BCE & KLD & NLL \\ \hline
        JSD & 0.4036 & \textbf{0.3407} & 0.3482 \\
        SWD ($\times10^{-5}$) & 9.843 & 9.958 & \textbf{7.407} \\
    \hline
    \end{tabular}
    \label{tab:scud_eva}
\end{table}

\section{Implementation}
\subsection{Motion Prediction via Deep Learning}
A U-Net architecture \cite{unet_2015} with the PELU output layer is utilized to generate one-shot motion predictions. The network consists of four downsampling layers and four corresponding upsampling layers with skip connections. A lite version with 32/64/128/256 filters is implemented. More detailed hyperparameters can be found in the configuration from the provided repository. The input $\bm{x}$ to the neural network is a stack of masks $I_m$, indicating the targeted object's positions, and a scene image $I_s$. The input can be regarded as a multi-channel image and each channel has the same size $W\times H$. The output is a stack of processed energy spaces, and each of them indicates the future position distribution at a specific time instant. The variance of the Gaussian distribution mask $\bm{A}$ is 10 pixels. The model is trained on four datasets. The first one is the Stanford Drone Dataset \cite{robicquet_2016_SDD} as shown in Fig. \ref{fig:sdd_experiment} for qualitative comparison, and the training process follows the same procedure as in \cite{ynet_2021}. The second one is a synthetic crossroad dataset for quantitative analysis, with 300 trajectories generated for training, as shown in Fig. \ref{fig:scud_experiment}. The other ones include a warehouse dataset \cite{ze_2023} and a hospital dataset for the validation of the proposed DOV pipeline. The warehouse synthetic dataset contains 580 trajectories that are collected in the environment with random motion noise and the background image resolution is $330\times293$.The hospital synthetic dataset contains 660 trajectories with larger motion uncertainty than the warehouse scene and the background image resolution is $321\times 321$. In synthetic datasets, the pedestrian model is omnidirectional and moves along predefined paths with random velocity noise. 

\subsection{Proactive Collision Avoidance}
As illustrated in Fig. \ref{fig:pipline}, given environmental data and the locations of AMRs and dynamic obstacles from a vision and tracking system, the multimodal motion predictor outputs stacked probability maps indicating future positions of the target object at each time step from 1 to $N$. 
After drawing samples from these probability maps, the CGF post-processing method is applied for each time step, producing a stack of geometric maps, with each map containing predicted potential ellipsoid obstacles. MPC then uses these stacked geometric maps to plan a collision-free trajectory for the AMR. The motion model for AMRs is the non-holonomic unicycle model as in \cite{ze_2023}.
The sampling time of the system is 0.2 s. The predictive horizon $N$ is 20, which means 4 s ahead. In MPC, the critical horizon $N_{\text{crit}}$ is 5. For the DOA term, both the model weight $\alpha_m$ and the time weight $\beta_m$ are pre-defined. The model weight is currently not used and set to be 1. The time weight is designed to decrease as the time step increases. Values of all weights can be found in the configuration file from the given code repository.

\section{Evaluation}
In this section, both qualitative and quantitative evaluations of the proposed method are introduced. Realistic simulation is presented via ROS 2 and Gazebo\footnote{\raggedright Code repository: {\tiny\url{https://github.com/Woodenonez/DyObAv_MPCnEBM_Warehouse}}}. Full video of different scenarios is available\footnote{\raggedright Full video: \url{https://youtu.be/DpeadFZgl-Y}}.

\subsection{Motion Prediction Evaluation}
In motion prediction tasks, the capture of diversity is emphasized to increase the likelihood of matching the ground-truth trajectory. For downstream planning, the focus is on ensuring the predictions are closely aligned with the ground-truth probability distribution.
As implied in Fig. \ref{fig:sdd_experiment}, training via NLL produces concentrated predictions. Since real-world datasets don't come with ground-truth distributions and it is difficult to estimate a distribution due to the lack of data, for quantitative analysis, a simulated crossroad scene is used as shown in Fig. \ref{fig:scud_experiment}, where a pedestrian may go straight, turn left or right at the intersection. 
To increase local uncertainty, the right side is an open area, and pedestrians randomly pick endpoints along the right edge of the map. The ground-truth distribution is estimated by fitting samples into a GMM (bottom left corner in Fig. \ref{fig:scud_experiment}) and compared to output probability maps from different models in terms of Jensen-Shannon Divergence (JSD) and Earth Mover's Distance (EMD, or the Wasserstein distance \cite{overcoming_2019}). 
JSD measures overall similarity rather than the actual distance between distribution peaks. EMD calculates the minimum ``effort" to transform one distribution into another, which is the product of the amount of moved probability mass and the corresponding distance. EMD depends on both the distance between distributions and their overall shapes. Given EMD's computational complexity in high dimensions, Sliced Wasserstein Distance (SWD) \cite{semd_2015} is chosen for Monte Carlo approximation. As shown in TABLE \ref{tab:scud_eva}, KLD and NLL perform similarly on JSD, while NLL outperforms others on SWD.
In our experiments, the original NLL loss performs adequately in the simple crossroad dataset. However, without the PELU and soft loss modifications, the training is unstable and prone to failure for real-world and complex datasets. The model also generates severe noise without the PELU layer, making it infeasible for downstream tasks.

\begin{figure}[t]
    \centering
    \includegraphics[width=0.8\linewidth]{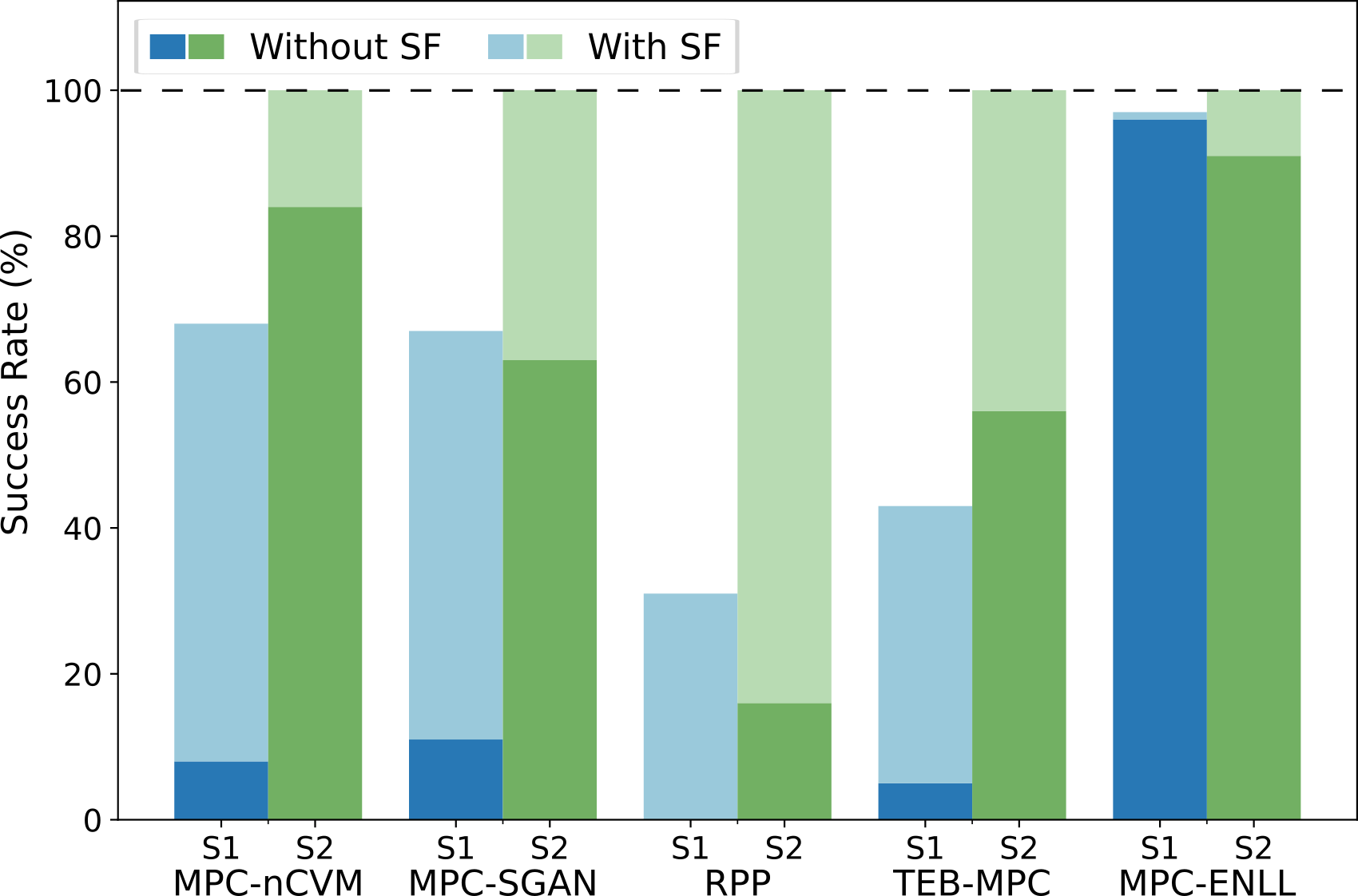}
    \vspace{-3mm}
    \caption{Success rates of different methods with and without the cooperative SF pedestrian model. S1 and S2 mean Scenario 1 and Scenario 2.}
    \label{fig:success_sf}
\end{figure}

\subsection{Obstacle Avoidance Evaluation}
The obstacle avoidance performance is evaluated by ensuring that AMRs avoid collisions with all obstacles. 
To thoroughly assess the proposed MPC-ENLL method, we compare various obstacle avoidance strategies, with and without cooperative pedestrian behavior, and then examine the impact of different predictors, including BCE (non-EBL), KLD (EBL), and ENLL (EBL). The effectiveness of the CGF grouping is also validated.
It is important to note that in all tests, motion noise is added to human movement.

\begin{figure}[t]
    \centering
    \includegraphics[width=0.7\linewidth]{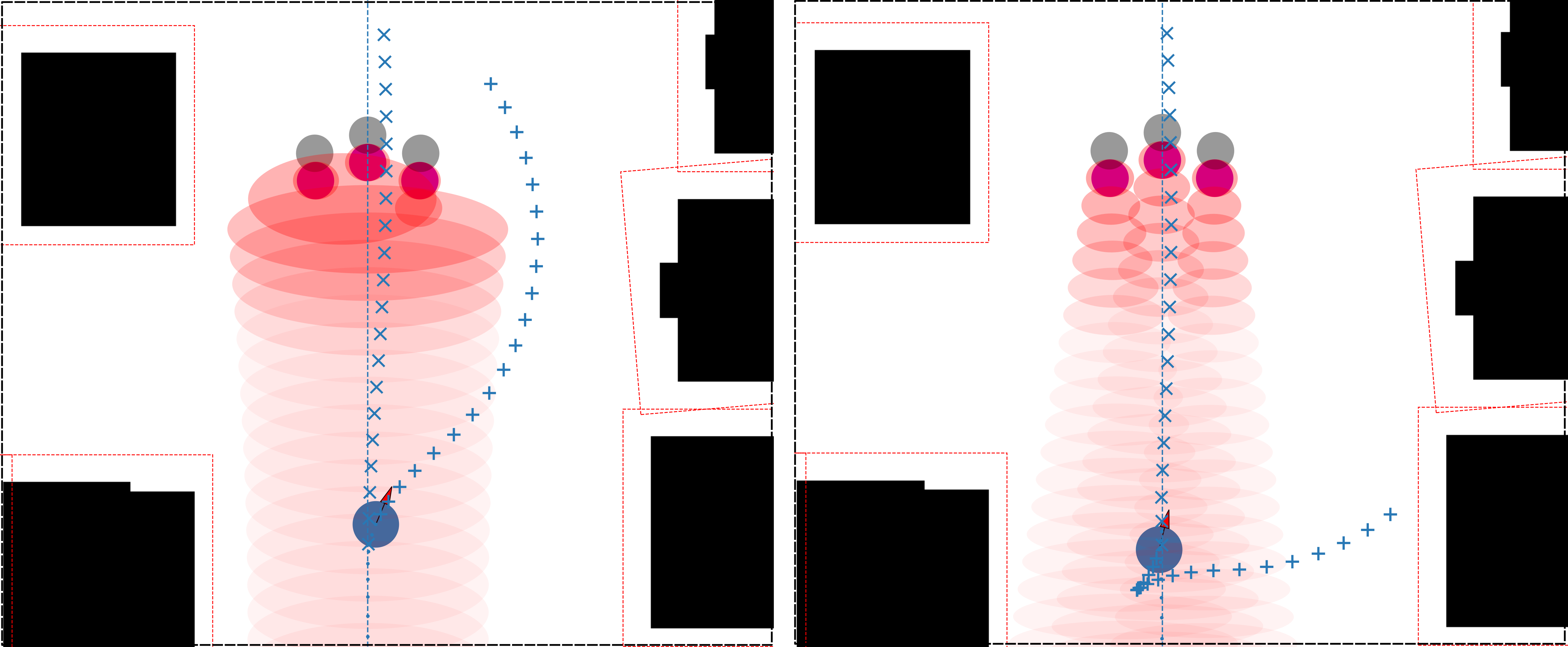}
    \vspace{-3mm}
    \caption{Comparison between grouped prediction (left) and individual prediction (right). As in Fig. \ref{fig:s3}, the blue circle is the AMR, and the magenta circles are pedestrians. Red ellipses are predicted positions of pedestrians.}
    \label{fig:group_comparison}
\end{figure}

Four test scenarios are set. Scenarios 1-3 are in a warehouse and Scenario 4 is in a hospital area, as shown in Fig. \ref{fig:s34_ros}. Scenarios 1 and 2 are inherited from \cite{ze_2023}. In Scenario 1, a pedestrian suddenly emerges from a corner, while in Scenario 2, a pedestrian abruptly changes direction in front of a robot. Scenario 3 involves two AMRs and four pedestrians meeting at an intersection, as in Fig. \ref{fig:s3}. In Scenario 4, a longer test is conducted, where an AMR operates in a hospital area and interacts with two pedestrians multiple times. The complete episode is available in the attached video. Multiple metrics are adopted, including linear and angular action smoothness (second derivative), clearance to obstacles, deviation from the reference path, solving time of motion planner (the duration from the receipt of motion prediction results and reference trajectories to the generation of the action), and success rate (defined as reaching the goal within a time limit without any collisions).
In addition to MPC-WTA and DWA-CVM methods \cite{ze_2023}, several baselines are evaluated: Regulated Pure Pursuit (RPP) \cite{rpp_2023} which is a stop-and-wait method, Timed-Elastic Band (TEB) \cite{teb_2015} which is a trajectory optimization method that generates collision-free reference trajectories for MPC, CVM with angular noise (nCVM) \cite{cvm_2020} integrated with MPC, and learning-based Social-GAN (SGAN) \cite{sgan_2018_CVPR} with MPC.
The results are in TABLE \ref{tab:eva1}. While learning-based control methods are not in the evaluation, as this study focuses on deterministic and optimization-based approaches, according to the study in \cite{ddpg_2024}, learning-based methods demonstrate superior performance in avoiding complex static obstacles but struggle with moving obstacles with limited training. The use of learning-based techniques is a potential future extension of this work.
As in TABLE \ref{tab:eva1}, our MPC-ENLL outperforms the other methods on success rate. Methods using the nCVM or SGAN predictor exhibit low success rates in Scenario 1 due to their failure to capture multimodal motion. Clearance is also notable. With proper motion prediction, MPC-ENLL maintains a large clearance from dynamic obstacles while utilizing space around static obstacles. 
For motion prediction, on an NVIDIA GTX 1650 Max-Q, the model inference time of the one-shot EBL predictor is about 3 ms per object, which is about 19 times faster than the WTA-based predictor (58 ms).

\begin{table*}[t]
    \centering
    % \captionsetup{justification=centering}
    \caption{Evaluation results of different methods (average over 100 runs)}
    \vspace{-2mm}
    \label{tab:eva1}
    \begin{tabular}{|c|l|c|c|c|c|c|c|c|c|c|c|}
    \hline
         \multirow{2}*{Scenario} & \multirow{2}*{Method} & \multicolumn{2}{c|}{Smoothness} & \multicolumn{2}{c|}{Clearance (m)}  & \multicolumn{3}{c|}{Deviation (m)} & \multicolumn{2}{c|}{Solving time (sec)} & \multirow{2}*{Success (\%)} \\
        \cline{3-11}
        & & linear & angular & static & dynamic & mean & std & max & mean & max & \\
        \cline{1-12}
        
        \multirow{6}*{Scenario 1} 
        %                         lin-s   ang-s   sta     dyn     dev     std     max     sol     max     suc
        & MPC-WTA* \cite{ze_2023}   & \textbf{0.030} & \textbf{0.039} & 0.501 & 0.786 & 0.434 & 0.267 & 1.095 & 0.050 & 0.125 & 94 \\
        & DWA-CVM* \cite{dwa_2019}   & 0.045 & 0.094 & 0.989 & 0.308 & 0.204 & 0.142 & 0.704 & 0.195 & 0.413 & 48 \\
        % & RPP \cite{rpp_2023}       & -&	-&	-&	-&	-&	-&	-&	-&	-&	0 \\
        % & TEB-MPC \cite{teb_2015} & 0.015&	0.009&	0.728&	0.312&	0.198&	0.109&	0.394&	0.005&	0.102&	5 \\
        % & MPC-nCVM \cite{cvm_2020}  & 0.063&	0.092&	0.594&	0.337&	0.856&	1.395&	6.009&	0.012&	0.102&	8 \\
        % & MPC-SGAN \cite{proactive_2023}	& 0.057&    0.062&	0.542&	0.341&	0.395&	0.243&	0.875&	0.012&	0.102&	11 \\
        & MPC-BCE	& 0.048&	0.050&	0.218&	0.612&	0.498&	0.301&	0.99&	0.012&	0.103&	57 \\
        & MPC-KLD	& 0.046&	0.054&	0.254&	0.515&	0.475&	0.289&	0.999&	0.010&	0.102&	43 \\
        & \textbf{MPC-ENLL}	& 0.040&	0.048&	0.333&	0.704&	0.632&	0.375&	1.270&	\textbf{0.008}&	\textbf{0.036}&	\textbf{96} \\
        
        \hline
        
        \multirow{7}*{Scenario 2} 
        %                         lin-s   ang-s   sta     dyn     dev     std     max     sol     max     suc
        & MPC-WTA*   & 0.046&	0.042&	0.574&	0.863&	0.224&	0.160&	1.064&	0.052&	0.121&	81 \\
        & DWA-CVM*   & 0.042&	0.034&	0.651&	0.298&	0.205&	0.109&	0.398&	0.207&	0.323&	62 \\
        & RPP \cite{rpp_2023}      & 0.027&	$\approx$0&	0.5&	0.393&	0.191&	0.109&	0.3&	$\approx$0&	$\approx$0&	16 \\
        & TEB-MPC \cite{teb_2015} & \textbf{0.010}&	\textbf{0.012}&	0.395&	0.474&	0.206&	0.111&	0.453&	\textbf{0.005}&	\textbf{0.102}&	56 \\
        & MPC-nCVM \cite{cvm_2020} & 0.046&	0.031&	0.475&	0.730&	0.229&	0.115&	0.500&	0.011&	0.103&	84 \\
        & MPC-SGAN \cite{proactive_2023}	& 0.031&	0.019&	0.472&	0.553&	0.243&	0.269&	2.893&	0.007&	0.102&	63 \\
        & MPC-BCE	& 0.049&	0.048&	0.303&	1.601&	0.391&	0.149&	0.859&	0.010&	0.103&	89 \\
        & MPC-KLD	& 0.058&	0.049&	0.366&	1.094&	0.343&	0.125&	0.574&	0.011&	0.103&	75 \\
        & \textbf{MPC-ENLL}	& 0.040&	0.024&	0.492&	0.927&	0.210&	0.109&	0.479&	0.010&	0.103&	\textbf{91} \\
        \hline
        
        \multirow{1}*{Scenario 3}
        & \textbf{MPC-ENLL}   & \textbf{0.027}&	\textbf{0.048}&	0.733&	0.898&	1.262&	0.941&	2.201&	\textbf{0.013}&	\textbf{0.103}&	\textbf{77} \\
        \hline
        
        \multirow{3}*{Scenario 4}
        & MPC-BCE   & 0.027& 0.033&   0.042&	0.731&	0.585&	0.397&	1.226&	0.037&	0.121&	19 \\
        & MPC-KLD   & 0.029& 0.030&   0.071&	0.662&	0.918&	0.708&	2.401&	0.027&	0.118&	18 \\
        & \textbf{MPC-ENLL}   & \textbf{0.017}&	\textbf{0.022}&	0.099&	0.586&	0.017&	0.350&	0.893&	\textbf{0.019}&	\textbf{0.109}&	\textbf{90} \\
        \hline
    \end{tabular}
    
    {\raggedright Methods in bold font are proposed in this work. Metrics in bold font are the best results. Starred (*) results are from \cite{ze_2023}. In all scenarios, results of methods with success rates lower than $15\%$ are omitted. Scenario 3 shows worse data in two robots. \par}
\end{table*}

In TABLE \ref{tab:eva1}, all pedestrians in Scenarios 1 and 2 are assumed to be non-cooperative and ignore AMRs, representing a worst-case scenario. To evaluate a more common condition with cooperative pedestrians and test the robustness, a modified Social Force (SF) model \cite{sf_1995} is applied, where pedestrians avoid AMRs actively.
As shown in Fig. \ref{fig:success_sf}, in Scenario 1, due to the pedestrian's abrupt appearance, even with the SF pedestrian model, other methods still struggle. In Scenario 2, all methods achieve perfect collision avoidance. 
These results highlight the difference in the collision-avoidance ability of different methods across scenarios, with the MPC-ENLL method demonstrating prominent performance regarding cooperative or non-cooperative dynamic obstacles. 
To compare different loss functions in network training, MPC-BCE and MPC-KLD are also tested as in TABLE \ref{tab:eva1}. In Scenarios 1, 3, and 4, they have lower success rates compared to the proposed MPC-ENLL method due to the overestimation of future occupied areas. In Scenario 2, with larger free space than the other scenarios, they achieve more comparable performance.

The FRP is considered when selecting the CGF grouping strategy. As illustrated in Fig. \ref{fig:group_comparison}, where three pedestrians are working towards an AMR, if the individual prediction is implemented, these predictions form a non-convex obstruction causing the robot to freeze. After grouping predictions via CGF grouping, the AMR regards the occupied area in its entirety and successfully avoids it.

% \subsection{Safety aspects}
In all MPC experiments, the solver time is restricted to 0.1 s for real-time performance, meaning that the solving process can be interrupted. This can lead to premature solutions without the collision-free guarantee \cite{ze_2023}. In production, a dedicated processor typically runs the solver that is further optimized. Additional safety layers including extra sensors are implemented to ensure safety. These measures are beyond the scope of this work and are not discussed.

\subsection{Long-term Robustness}
Apart from Scenario 4, to further examine the robustness of the proposed pipeline, long-term running experiments in the warehouse (with cooperative SF pedestrians) are implemented.
To increase the uncertainty and interaction frequency, three AMRs and three pedestrians operating in close proximity are included, set to follow predefined routes. The location and timing of interactions between AMRs and pedestrians varied randomly due to the human motion noise. In three trials, each AMR encountered over 20 pedestrian interactions (defined as instances where a pedestrian influenced the robot’s actions) and multiple fleet collision avoidance cases. 
The proposed approach demonstrates robustness and adaptability throughout the tests. Robots autonomously select different strategies based on the situation, such as taking detours when necessary, cutting shorter paths if safe, reversing in response to dynamic obstacles, etc. 
With distributed MPC, collision avoidance between multiple AMRs is also handled smoothly.

\subsection{Simulation in Gazebo}
\begin{figure}
    \centering
    \includegraphics[width=0.7\linewidth]{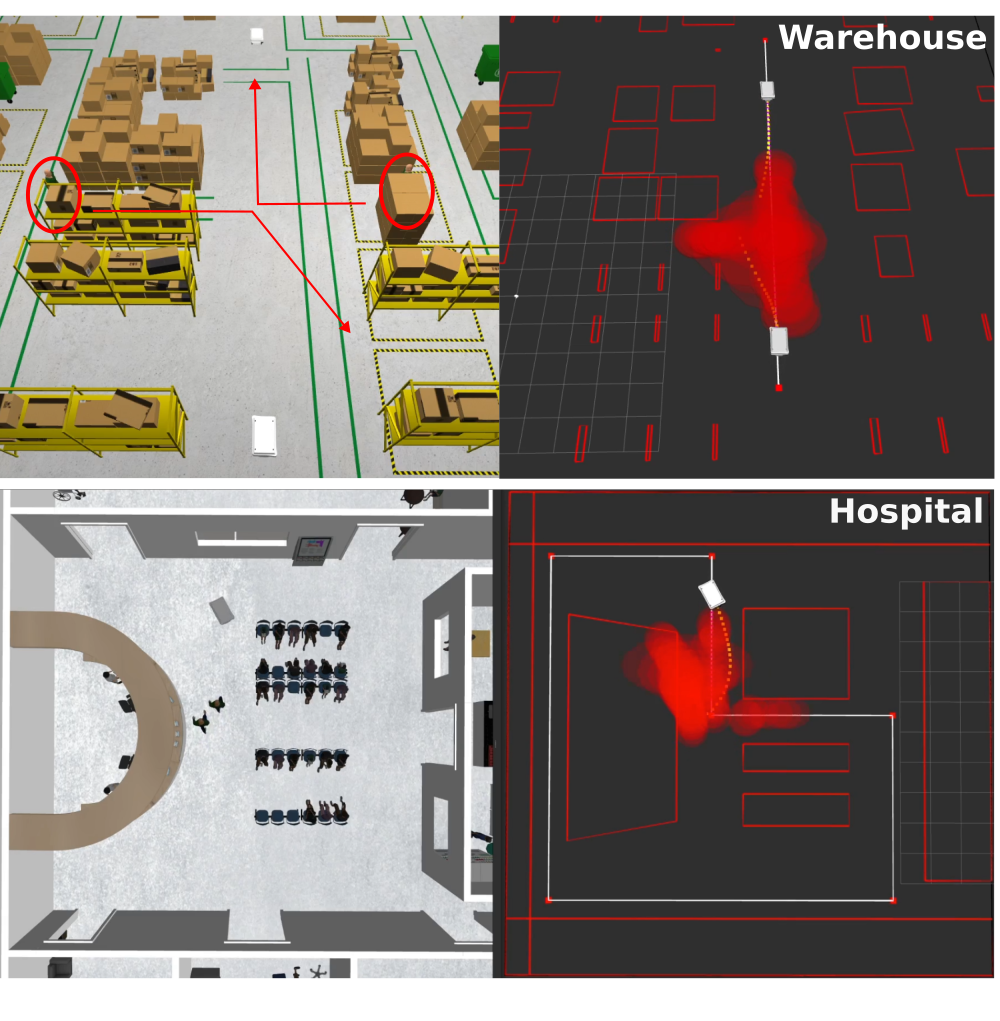}
    \vspace{-5mm}
    \caption{Gazebo simulation of Scenarios 3 (up) and 4 (down). Red circles and arrows indicate the initial positions and intended routes of pedestrians. In RViz visualization, red ellipses are predictions from the energy-based predictor and yellow dotted lines are planned trajectories by MPC.}
    \label{fig:s34_ros}
\end{figure}

The proposed pipeline is deployed in Gazebo for realistic simulation. Fig. \ref{fig:s34_ros} demonstrates the warehouse and hospital environments. The corresponding code and environments are available in the provided repository.

\subsection{Relevant Hyperparameters and Normalized Masks}
In this session, we briefly discuss some relevant hyperparameters and analyze the use of normalized masks. In the original Gibbs-Boltzmann distribution, there is an inverse temperature parameter scaling energy values in the numerator of \eqref{eq:boltzman}, which is normally 1 as in this work. In practice, this can be used to tune the conservativeness of the prediction. Another parameter that affects the conservatives of outputs is the variance of the ground-truth Gaussian mask $\bm{A}$. BCE and KLD are more sensitive to the variance of the mask, as they attempt to match it exactly. The proposed ENLL loss is less affected by the mask width, offering greater robustness.

In different loss functions, different forms of masks are employed, either the estimated ground-truth probability map $\bm{A}$ or its normalization $\bar{\bm{A}}$. 
For the BCE loss, normalization is necessary for mathematical soundness since it requires a Sigmoid output layer. Otherwise, it becomes inapplicable. 
For the KLD loss, as defined in \eqref{eq:loss_kld}, normalizing the mask scales it by the reciprocal of the maximum value in $\bm{A}$. When evaluating using the setup in TABLE \ref{tab:scud_eva}, with normalized masks, the JSD is $0.3267$ and the SWD is $9.6767\times10^{-5}$, which are slightly better than KLD without normalized masks but not significantly. Moreover, normalization had no essential impact on obstacle avoidance performance as it cannot solve the problem of over-conservativeness. 
For the ENLL loss, using the normalized mask scales up its negative component, which may bias training toward stronger alignment with the mask. In practice, ENLL trained with unnormalized masks produced slightly more concentrated predictions but did not lead to significant differences in obstacle avoidance performance.

\section{Conclusion and Future Work}
In this study, we proposed a new energy-based learning method for multimodal motion prediction and extended the MPC formulation for fleet coordination in an integrated collision-free navigation pipeline for autonomous mobile robots. 
The motion predictor based on energy-based learning is shown to be more suitable for downstream motion planning tasks. The combined MPC-ENLL approach outperforms other popular obstacle avoidance approaches with or without the cooperative pedestrian model. This pipeline is also implemented in ROS2 and simulated in Gazebo under a warehouse environment, which shows feasibility and safety. 

One limitation of this work is the use of simulated environments, with complex static–dynamic obstacle interactions left for future investigation.
Other future work could include three aspects: extending the pipeline to include the vision and tracking system; implementing a reference generator for MPC to achieve fast convergence; and considering industrial transport tasks involving more mobile robots and task schedulers.

% \section*{Acknowledgments}
% This should be a simple paragraph before the References to thank those individuals and institutions who have supported your work on this article.

\bibliographystyle{IEEEtran}
\bibliography{IEEEabrv, main}

\begin{thebibliography}{10}
\providecommand{\url}[1]{#1}
\csname url@rmstyle\endcsname
\providecommand{\newblock}{\relax}
\providecommand{\bibinfo}[2]{#2}
\providecommand\BIBentrySTDinterwordspacing{\spaceskip=0pt\relax}
\providecommand\BIBentryALTinterwordstretchfactor{4}
\providecommand\BIBentryALTinterwordspacing{\spaceskip=\fontdimen2\font plus
\BIBentryALTinterwordstretchfactor\fontdimen3\font minus \fontdimen4\font\relax}
\providecommand\BIBforeignlanguage[2]{{%
\expandafter\ifx\csname l@#1\endcsname\relax
\typeout{** WARNING: IEEEtran.bst: No hyphenation pattern has been}%
\typeout{** loaded for the language `#1'. Using the pattern for}%
\typeout{** the default language instead.}%
\else
\language=\csname l@#1\endcsname
\fi
#2}}

\bibitem{agv_2020}
M.~{De Ryck}, M.~Versteyhe, and F.~Debrouwere, ``Automated guided vehicle systems, state-of-the-art control algorithms and techniques,'' \emph{Journal of Manufacturing Systems}, vol.~54, pp. 152--173, 2020.

\bibitem{frp_2010}
P.~Trautman and A.~Krause, ``Unfreezing the robot: Navigation in dense, interacting crowds,'' in \emph{IROS}, 2010, pp. 797--803.

\bibitem{unfrozen_2020}
A.~J. Sathyamoorthy, U.~Patel, T.~Guan, and D.~Manocha, ``Frozone: Freezing-free, pedestrian-friendly navigation in human crowds,'' \emph{Robotics and Automation Letters}, vol.~5, no.~3, pp. 4352--4359, 2020.

\bibitem{attentionbased_2023}
S.~Liu, P.~Chang, Z.~Huang, N.~Chakraborty, K.~Hong, W.~Liang, D.~L. McPherson, J.~Geng, and K.~Driggs-Campbell, ``Intention aware robot crowd navigation with attention-based interaction graph,'' in \emph{ICRA}, 2023, pp. 12\,015--12\,021.

\bibitem{socialnavi_2023}
C.~Mavrogiannis, F.~Baldini, A.~Wang, D.~Zhao, P.~Trautman, A.~Steinfeld, and J.~Oh, ``Core challenges of social robot navigation: A survey,'' \emph{ACM Transactions on Human-Robot Interaction}, vol.~12, no.~3, 2023.

\bibitem{dynchannel_2019}
C.~Cao, P.~Trautman, and S.~Iba, ``Dynamic channel: A planning framework for crowd navigation,'' in \emph{ICRA}, 2019, pp. 5551--5557.

\bibitem{dynsys_2022}
L.~Huber, J.-J. Slotine, and A.~Billard, ``Avoiding dense and dynamic obstacles in enclosed spaces: Application to moving in crowds,'' \emph{IEEE Transactions on Robotics}, vol.~38, no.~5, pp. 3113--3132, 2022.

\bibitem{due_2012}
N.~E. Du~Toit and J.~W. Burdick, ``Robot motion planning in dynamic, uncertain environments,'' \emph{IEEE Transactions on Robotics}, vol.~28, no.~1, pp. 101--115, 2012.

\bibitem{proactive_2023}
L.~Heuer, L.~Palmieri, A.~Rudenko, A.~Mannucci, M.~Magnusson, and K.~O. Arras, ``Proactive model predictive control with multi-modal human motion prediction in cluttered dynamic environments,'' in \emph{IROS}, 2023, pp. 229--236.

\bibitem{groupbased_2022}
A.~Wang, C.~Mavrogiannis, and A.~Steinfeld, ``Group-based motion prediction for navigation in crowded environments,'' in \emph{Conference on Robot Learning (CoRL)}, 2022, pp. 871--882.

\bibitem{ze_2023}
Z.~Zhang, H.~Hajieghrary, E.~Dean, and K.~Åkesson, ``Prescient collision-free navigation of mobile robots with iterative multimodal motion prediction of dynamic obstacles,'' \emph{Robotics and Automation Letters}, vol.~8, no.~9, pp. 5488--5495, 2023.

\bibitem{cvm_2020}
C.~Schöller, V.~Aravantinos, F.~Lay, and A.~Knoll, ``What the constant velocity model can teach us about pedestrian motion prediction,'' \emph{Robotics and Automation Letters}, vol.~5, no.~2, pp. 1696--1703, 2020.

\bibitem{sf_1995}
D.~Helbing and P.~Molnar, ``Social force model for pedestrian dynamics,'' \emph{Physical review E}, vol.~51, no.~5, p. 4282, 1995.

\bibitem{rvo_2008}
J.~van~den Berg, M.~Lin, and D.~Manocha, ``Reciprocal velocity obstacles for real-time multi-agent navigation,'' in \emph{ICRA}, 2008, pp. 1928--1935.

\bibitem{slstm_2016_CVPR}
A.~Alahi, K.~Goel, V.~Ramanathan, A.~Robicquet, L.~Fei-Fei, and S.~Savarese, ``Social {LSTM}: Human trajectory prediction in crowded spaces,'' in \emph{CVPR}, June 2016.

\bibitem{sgan_2018_CVPR}
A.~Gupta, J.~Johnson, L.~Fei-Fei, S.~Savarese, and A.~Alahi, ``Social {GAN}: Socially acceptable trajectories with generative adversarial networks,'' in \emph{CVPR}, June 2018.

\bibitem{overcoming_2019}
O.~Makansi, E.~Ilg, O.~Cicek, and T.~Brox, ``Overcoming limitations of mixture density networks: A sampling and fitting framework for multimodal future prediction,'' in \emph{CVPR}, 2019.

\bibitem{ynet_2021}
K.~Mangalam, Y.~An, H.~Girase, and J.~Malik, ``From goals, waypoints \& paths to long term human trajectory forecasting,'' in \emph{ICCV}, 2021, pp. 15\,213--15\,222.

\bibitem{mpssd_2020}
D.~Ridel, N.~Deo, D.~Wolf, and M.~Trivedi, ``Scene compliant trajectory forecast with agent-centric spatio-temporal grids,'' \emph{Robotics and Automation Letters}, vol.~5, no.~2, pp. 2816--2823, 2020.

\bibitem{trajectronpp_2020}
T.~Salzmann, B.~Ivanovic, P.~Chakravarty, and M.~Pavone, ``Trajectron++: Dynamically-feasible trajectory forecasting with heterogeneous data,'' in \emph{European Conference on Computer Vision (ECCV)}.\hskip 1em plus 0.5em minus 0.4em\relax Springer, 2020, pp. 683--700.

\bibitem{end2end_2022}
K.~Guo, W.~Liu, and J.~Pan, ``End-to-end trajectory distribution prediction based on occupancy grid maps,'' in \emph{CVPR}, 2022, pp. 2232--2241.

\bibitem{ddpg_2024}
K.~Ceder, Z.~Zhang, A.~Burman, I.~Kuangaliyev, K.~Mattsson, G.~Nyman, A.~Petersén, L.~Wisell, and K.~Åkesson, ``Bird’s-eye-view trajectory planning of multiple robots using continuous deep reinforcement learning and model predictive control,'' in \emph{IROS}, 2024, pp. 8002--8008.

\bibitem{ze_2021}
Z.~Zhang, E.~Dean, Y.~Karayiannidis, and K.~Åkesson, ``Motion prediction based on multiple futures for dynamic obstacle avoidance of mobile robots,'' in \emph{CASE}, 2021, pp. 475--481.

\bibitem{dist_2022}
F.~Bertilsson, M.~Gordon, J.~Hansson, D.~Möller, D.~Söderberg, Z.~Zhang, and K.~Åkesson, ``Centralized versus distributed nonlinear model predictive control for online robot fleet trajectory planning,'' in \emph{CASE}, 2022, pp. 701--706.

\bibitem{robicquet_2016_SDD}
A.~Robicquet, A.~Sadeghian, A.~Alahi, and S.~Savarese, ``Learning social etiquette: Human trajectory understanding in crowded scenes,'' in \emph{European Conference on Computer Vision (ECCV)}.\hskip 1em plus 0.5em minus 0.4em\relax Springer, 2016.

\bibitem{lecun_2006}
Y.~LeCun, S.~Chopra, R.~Hadsell, M.~Ranzato, F.~Huang, \emph{et~al.}, ``A tutorial on energy-based learning,'' \emph{Predicting structured data}, vol.~1, no.~0, 2006.

\bibitem{ebm_2020}
F.~K. Gustafsson, M.~Danelljan, G.~Bhat, and T.~B. Sch{\"o}n, ``Energy-based models for deep probabilistic regression,'' in \emph{European Conference on Computer Vision (ECCV)}.\hskip 1em plus 0.5em minus 0.4em\relax Springer, 2020, pp. 325--343.

\bibitem{ebmtraining_2021}
Y.~Song and D.~P. Kingma, ``How to train your energy-based models,'' \emph{arXiv preprint arXiv:2101.03288}, 2021.

\bibitem{elu_2016}
S.~H. Djork-Arné~Clevert, Thomas~Unterthiner, ``Fast and accurate deep network learning by exponential linear units ({ELU}s),'' in \emph{International Conference on Learning Representations (ICLR)}, 2016.

\bibitem{unet_2015}
O.~Ronneberger, P.~Fischer, and T.~Brox, ``U-{N}et: Convolutional networks for biomedical image segmentation,'' in \emph{Medical Image Computing and Computer-Assisted Intervention (MICCAI)}.\hskip 1em plus 0.5em minus 0.4em\relax Springer, 2015, pp. 234--241.

\bibitem{semd_2015}
N.~Bonneel, J.~Rabin, G.~Peyr{\'e}, and H.~Pfister, ``Sliced and radon wasserstein barycenters of measures,'' \emph{Journal of Mathematical Imaging and Vision}, vol.~51, pp. 22--45, 2015.

\bibitem{rpp_2023}
S.~Macenski, S.~Singh, F.~Martín, and J.~Ginés, ``Regulated pure pursuit for robot path tracking,'' \emph{Autonomous Robots}, vol.~47, p. 685–694, 2023.

\bibitem{teb_2015}
C.~Rösmann, F.~Hoffmann, and T.~Bertram, ``Timed-elastic-bands for time-optimal point-to-point nonlinear model predictive control,'' in \emph{ECC}, 2015, pp. 3352--3357.

\bibitem{dwa_2019}
M.~Missura and M.~Bennewitz, ``Predictive collision avoidance for the dynamic window approach,'' in \emph{ICRA}, 2019, pp. 8620--8626.

\end{thebibliography}

% \newpage

% \section{Biography Section}
% If you have an EPS/PDF photo (graphicx package needed), extra braces are
%  needed around the contents of the optional argument to biography to prevent
%  the LaTeX parser from getting confused when it sees the complicated
%  $\backslash${\tt{includegraphics}} command within an optional argument. (You can create
%  your own custom macro containing the $\backslash${\tt{includegraphics}} command to make things
%  simpler here.)
 
% \vspace{11pt}

% \bf{If you include a photo:}\vspace{-33pt}
% \begin{IEEEbiography}[{\includegraphics[width=1in,height=1.25in,clip,keepaspectratio]{fig1}}]{Michael Shell}
% Use $\backslash${\tt{begin\{IEEEbiography\}}} and then for the 1st argument use $\backslash${\tt{includegraphics}} to declare and link the author photo.
% Use the author name as the 3rd argument followed by the biography text.
% \end{IEEEbiography}

% \vspace{11pt}

% \bf{If you will not include a photo:}\vspace{-33pt}
% \begin{IEEEbiographynophoto}{John Doe}
% Use $\backslash${\tt{begin\{IEEEbiographynophoto\}}} and the author name as the argument followed by the biography text.
% \end{IEEEbiographynophoto}

\vfill

\end{document}